\pdfoutput=1
\documentclass[11pt,a4paper]{article}
\usepackage[hyperref]{emnlp2018}
\usepackage{times}

\usepackage{url}

\usepackage[english]{babel}
\usepackage{CJKutf8}
\usepackage{amsmath}
\usepackage{amssymb}
\usepackage[english]{babel}
\usepackage{graphicx}
\usepackage{multirow}
\usepackage{bm}
\usepackage{subcaption}
\DeclareCaptionFont{10pt}{\fontsize{10pt}{12pt}\selectfont}
\usepackage{booktabs}

\newcommand{\R}{\mathbb{R}}

\makeatletter
\newcommand{\ssymbol}[1]{$^{\@fnsymbol{#1}}$}
\makeatother
\let\vec\bm

\aclfinalcopy 

\title{simNet: Stepwise Image-Topic Merging Network for \\Generating Detailed and  Comprehensive  Image Captions}

\author{Fenglin Liu\textsuperscript{1}\thanks{\ \ Equal Contributions}, Xuancheng Ren\textsuperscript{2}\footnotemark[1], Yuanxin Liu\textsuperscript{1}, Houfeng Wang\textsuperscript{2} and Xu Sun\textsuperscript{2}\\
\textsuperscript{1}School of ICE, Beijing University of Posts and Telecommunications\\
\textsuperscript{2}MOE Key Laboratory of Computational Linguistics, School of EECS, Peking University\\
{\tt lfl@bupt.edu.cn, renxc@pku.edu.cn, yuanxinLIU@bupt.edu.cn}\\
{\tt \{wanghf, xusun\}@pku.edu.cn }\\
}

\date{}

\begin{document}

\maketitle

\begin{abstract}

The encode-decoder framework has shown recent success in image captioning. Visual attention, which is good at detailedness, and semantic attention, which is good at comprehensiveness, have been separately proposed to ground the caption on the image. In this paper, we propose the \textit{S}tepwise \textit{I}mage-Topic \textit{M}erging \textit{N}etwork (\textit{simNet}) that makes use of the two kinds of attention at the same time. At each time step when generating the caption, the decoder adaptively merges the attentive information in the extracted topics and the image according to the generated context, so that the visual information and the semantic information can be effectively combined. The proposed approach is evaluated on two benchmark datasets and reaches the state-of-the-art performances.\footnote{\ The code is available at \url{https://github.com/lancopku/simNet}}
\end{abstract}

\setlength{\abovedisplayskip}{3pt}
\setlength{\abovedisplayshortskip}{3pt}
\setlength{\belowdisplayskip}{3pt}
\setlength{\belowdisplayshortskip}{3pt}

\section{Introduction}

Image captioning attracts considerable attention in both natural language processing and computer vision. The task aims to generate a description in natural language grounded on the input image. It is a very challenging yet interesting task. On the one hand, it has to identify the objects in the image, associate the objects, and express them in a fluent sentence, each of which is a difficult subtask. On the other hand, it combines two important fields in artificial intelligence, namely, natural language processing and computer vision. More importantly, it has a wide range of applications, including text-based image retrieval, helping visually impaired people see \cite{wu2017automatic}, human-robot interaction \cite{das2017visual}, etc.

\begin{figure}[t]

\centering
\footnotesize

\begin{tabular}{@{}c p{0.23\textwidth}@{}}

\multirow{3}{*}{\includegraphics[width=0.25\textwidth]{ex.jpg}} & \textbf{Soft-Attention}: a open laptop computer sitting on top of a table \\
& \textbf{ATT-FCN}: a dog sitting on a desk with a laptop computer and mouse\\
& \textbf{simNet}: a open laptop computer and mouse sitting on a table with a dog nearby\\

\end{tabular}

\caption{Examples of using different attention mechanisms. 
Soft-Attention \cite{xu2015show} is based on visual attention. The generated caption is \textbf{detailed} in that it knows the visual attributes well (e.g. \textit{open}). However, it omits many objects (e.g. \textit{mouse} and \textit{dog}). ATT-FCN \cite{you2016image} is based on semantic attention. The generated caption is more \textbf{comprehensive} in that it includes more objects. However, it is bad at associating details with the objects (e.g. missing \textit{open} and mislocating \textit{dog}). simNet is our proposal that effectively merges the two kinds of attention and generates a detailed and comprehensive caption.}
\label{fig:ex-intro}
\end{figure}

Models based on the encoder-decoder framework have shown success in image captioning. According to the pivot representation, they can be roughly categorized into models based on visual information \cite{vinyals2015show,chen2015mind,mao2014deep,karpathy2014deep,karpathy2017deep}, and models based on conceptual information \cite{fang2015captions,you2016image,wu2016what}. The later explicitly provides the visual words (e.g. \textit{dog}, \textit{sit}, \textit{red}) to the decoder instead of the image features, and is more effective in image captioning according to the evaluation on benchmark datasets. However, the models based on conceptual information have a major drawback that it is hard for the model to associate the details
with the specific objects in the image, because the visual words are inherently unordered in semantics. Figure~\ref{fig:ex-intro} shows an example. For semantic attention, although \textit{open} is provided as a visual word, due to the insufficient use of visual information, the model gets confused about what objects \textit{open} should be associated with and thus discards \textit{open} in the caption. The model may even associate the details incorrectly, which is the case for the position of the dog. In contrast, models based on the visual information often are accurate in details but have difficulty in describing the image comprehensively and tend to only describe a subregion. 

\begin{figure}[t]

\centering
\includegraphics[width=0.46\textwidth]{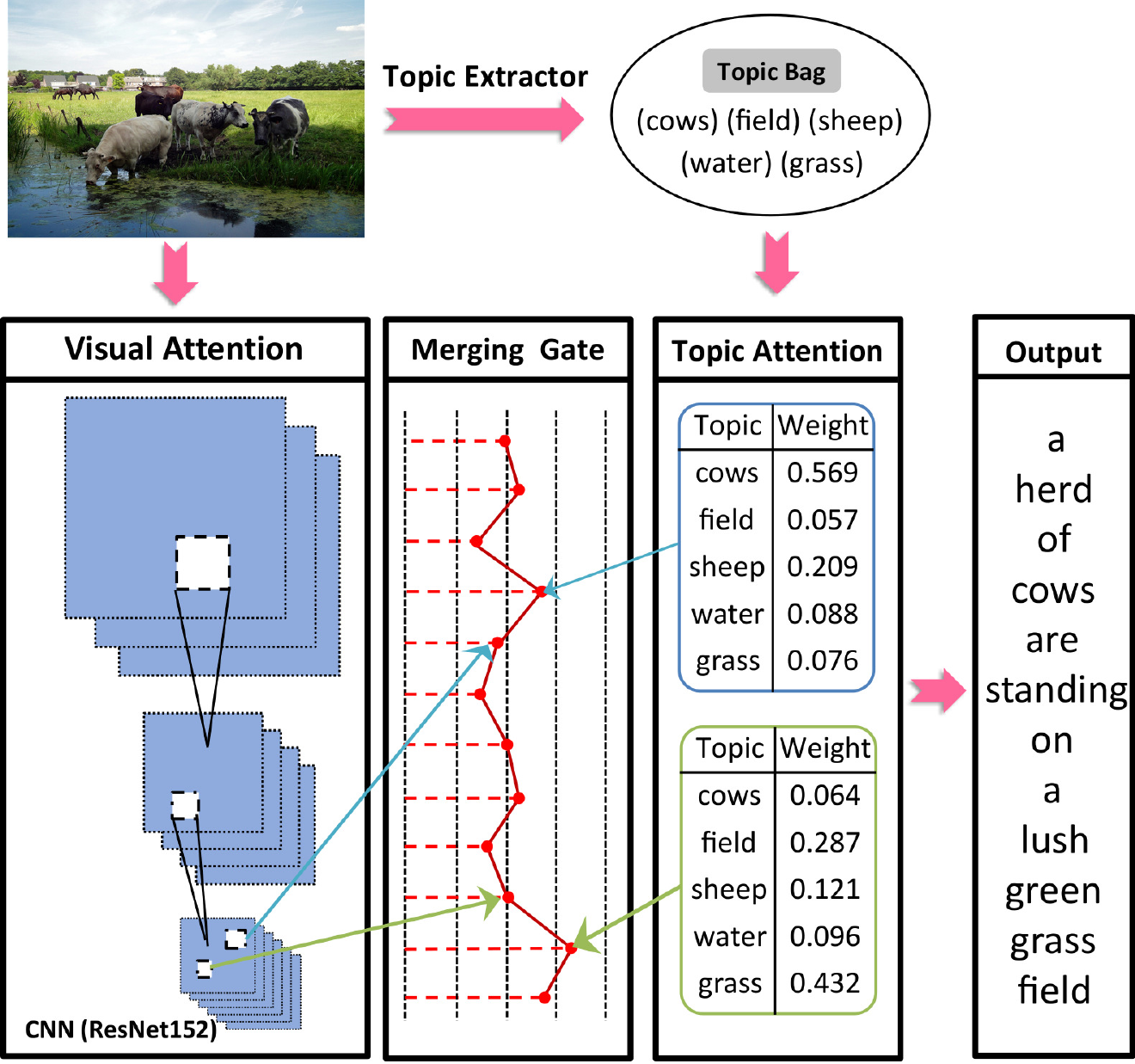}

\caption{Illustration of the main idea. The visual information captured by CNN and the conceptual information in the extracted topics are first condensed by attention mechanisms respectively. The merging gate then adaptively adjusts the weight between the visual information and the conceptual information for generating the caption.}
\label{fig:ex-running}
\end{figure}

In this work, we get the best of both worlds and integrate visual attention and semantic attention for generating captions that are both detailed and comprehensive. We propose a \textbf{Stepwise Image-Topic Merging Network} as the decoder to guide the information flow between the image and the extracted topics. At each time step, the decoder first extracts focal information from the image. Then, it decides which topics are most probable for the time step. Finally, it attends differently to the visual information and the conceptual information to generate the output word. Hence, the model can efficiently merge the two kinds of information, leading to outstanding results in image captioning. 

Overall, the main contributions of this work are:
\begin{itemize}

\item We propose a novel approach that can effectively merge the information in the image and the topics to generate cohesive captions that are both detailed and comprehensive. We refine and combine two previous competing attention mechanisms, namely visual attention and semantic attention, with an importance-based merging gate that effectively combines and balances the two kinds of information.

\item The proposed approach outperforms the state-of-the-art methods substantially on two benchmark datasets, Flickr30k and COCO, in terms of SPICE, which correlates the best with human judgments. Systematic analysis shows that the merging gate contributes the most to the overall improvement.

\end{itemize}

\begin{figure*}[t]

\centering
\subcaptionbox{The overall framework.\label{fig:overall}}{\includegraphics[height=6.5\baselineskip]{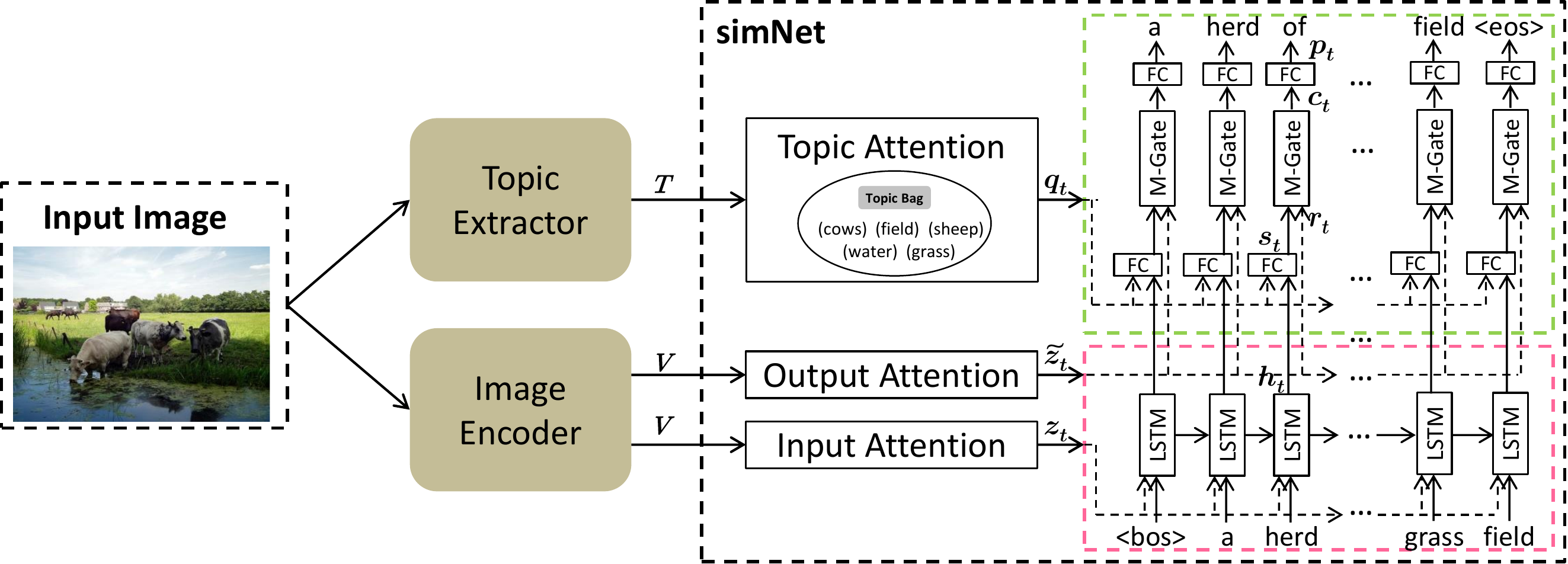}}
\hfil
\subcaptionbox{The data flow in the proposed simNet.\label{fig:tmn}}{\includegraphics[height=6.5\baselineskip]{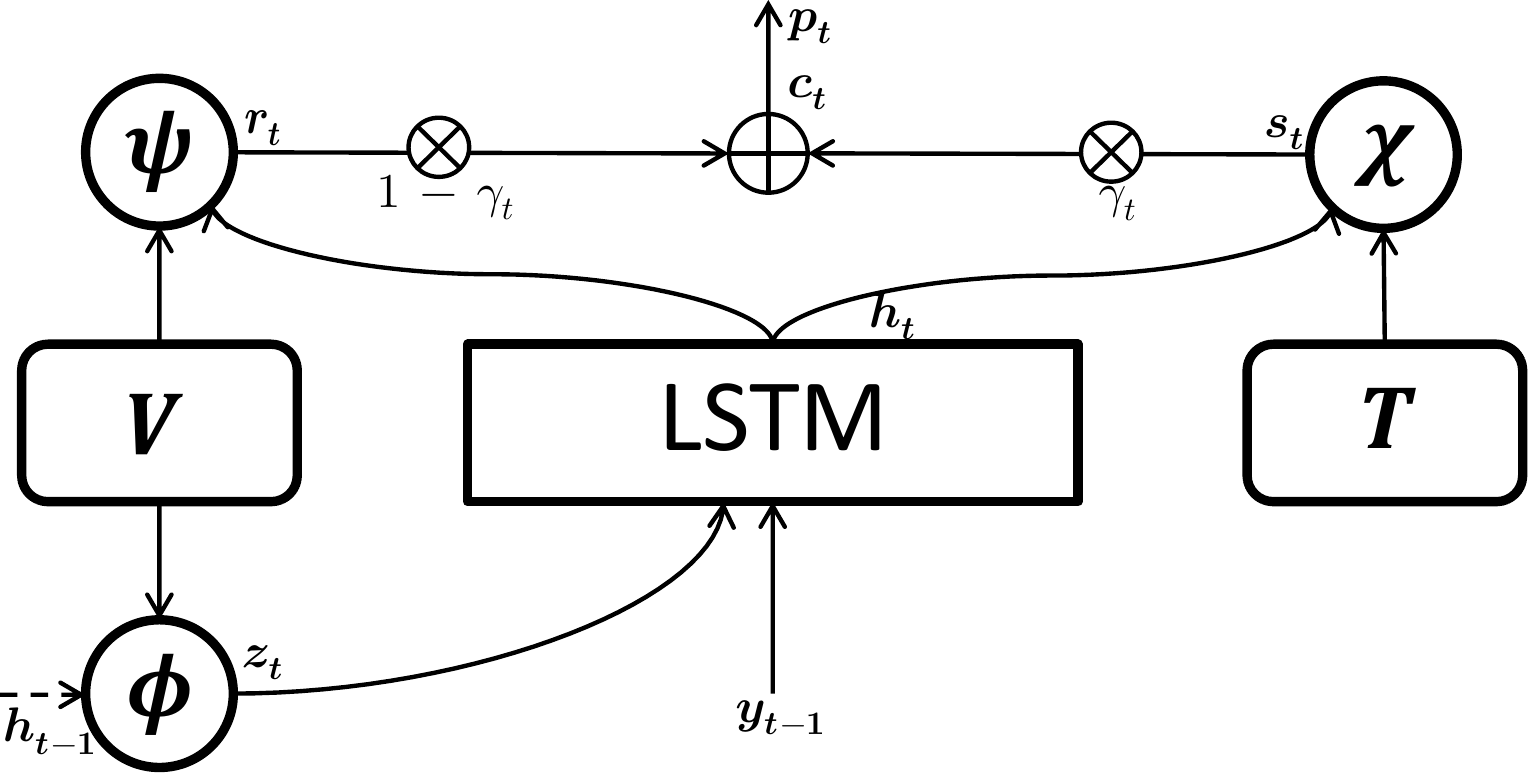}}

\caption{Illustration of the proposed approach. In the right plot, we use $\phi, \psi,\chi$ to denote input attention, output attention, and topic attention, respectively.}
\label{fig:model}
\end{figure*}

\section{Related Work}

A large number of systems have been proposed for image captioning. Neural models based on the encoder-decoder framework have been attracting increased attention in the last few years in several multi-discipline tasks, such as neural image/video captioning (NIC) and visual question answering (VQA) \cite{vinyals2015show,karpathy2014deep,venugopalan2015sequence,zhao2016partial,zhang2017visual}. State-of-the-art neural approaches \cite{anderson2018bottom,liu2018show,lu2018neural} incorporate the attention mechanism in machine translation \cite{bahdanau2014neural} to generate grounded image captions. Based on what they attend to, the models can be categorized into visual attention models and semantic attention models.

Visual attention models pay attention to the image features generated by CNNs. CNNs are typically pre-trained on the image recognition task to extract general visual signals \cite{xu2015show,chen2017scacnn,lu2017knowing}. The visual attention is expected to find the most relevant image regions in generating the caption. Most recently, image features based on predicted bounding boxes are used  \cite{anderson2018bottom,lu2018neural}. The advantages are that the attention no longer needs to find the relevant generic regions by itself but instead find relevant bounding boxes that are object orientated and can serve as semantic guides. However, the drawback is that predicting bounding boxes is difficult, which requires large datasets \cite{krishna2017visual} and complex models \cite{ren2015faster,ren2017faster}.

Semantic attention models pay attention to a predicted set of semantic concepts \cite{fang2015captions,you2016image,wu2016what}. The semantic concepts are the most frequent words in the captions, and the extractor can be trained using various methods but typically is only trained on the given image captioning dataset. This kind of approach can be seen as the extension of the earlier template-based slotting-filling approaches \cite{farhadi2010every,kulkarni2013babytalk}.

However, few work studies how to combine the two kinds of attention models to take advantage of both of them. 
On the one hand, due to the limited number of visual features, it is hard to provide comprehensive information to the decoder. On the other hand, the extracted semantic concepts are unordered, making it hard for the decoder to portray the details of the objects correctly.


This work focuses on combining the visual attention and the semantic attention efficiently to address their drawbacks and make use of their merits. The visual attention is designed to focus on the attributes and the relationships of the objects, while the semantic attention only includes words that are objects so that the extracted topics could be more accurate. The combination is controlled by the importance-based merging mechanism that decides at each time step which kind of information should be relied on. The goal is to generate image captions that are both detailed and comprehensive.

\nocite{lin18deconvolution,shuming2018,jingjing2018Upaired,xu2018emnlp}

\section{Approach}

Our proposed model consists of an image encoder, a topic extractor, and a stepwise merging decoder. Figure \ref{fig:model} shows a sketch. We first briefly introduce the image encoder and the topic extractor. Then, we introduce the proposed stepwise image-topic merging decoder in detail.

\subsection{Image Encoder}

For an input image, the image encoder expresses the image as a series of visual feature vectors $\vec{V}=\{\vec{v}_1, \vec{v}_2, \ldots, \vec{v}_k\}, \vec{v}_i \in \R^g$. Each feature corresponds to a different perspective of the image. The visual features serve as descriptive guides of the objects in the image for the decoder. We use a ResNet152 \cite{he2016deep}, which is commonly used in image captioning, to generate the visual features. The output of the last convolutional layer is used as the visual information:
\begin{align}
\vec{V} = \vec{W}^{V,I} \text{CNN}(\vec{I})
\end{align}
where $\vec{I}$ is the input image, and $\vec{W}^{V,I}$ shrinks the last dimension of the output.\footnote{For conciseness, all the bias terms of linear transformations in this paper are omitted.}

\subsection{Topic Extractor}

Typically, identifying an object requires a combination of visual features, and considering the limited capacity of the visual features, it is hard for the conventional decoder to describe the objects in the image comprehensively. An advance in image captioning is to provide the decoder with the semantic concepts in the image directly so that the decoder is equipped with an overall perspective of the image. The semantic concepts can be objects (e.g. \textit{person}, \textit{car}), attributes (e.g. \textit{off}, \textit{electric}), and relationships (e.g. \textit{using}, \textit{sitting}). We only use the words that are objects in this work, the reason of which is explained later. We call such words \textbf{topics}. The topic extractor concludes a list of candidate topic embeddings $\vec{T} = \{\vec{w}_1, \vec{w}_2, \ldots, \vec{w}_m\}, \vec{w}_i \in \R^e$ from the image, where $e$ is the dimension of the topic word embeddings. Following common practice \cite{fang2015captions,you2016image}, we adopt the weakly-supervised approach of Multiple Instance Learning \cite{zhang2006multiple} to build a topic extractor. Due to limited space, please refer to \citet{fang2015captions} for detailed explanation. 

Different from existing work that uses all the most frequent words in the captions as valid semantic concepts or visual words, we only include the object words (nouns) in the topic word list. Existing work relies on attribute words and relationship words to provide visual information to the decoder. However, it not only complicates the extracting procedure but also contributes little to the generation. For an image containing many objects, the decoder is likely to combine the attributes with the objects arbitrarily, as such words are specific to certain objects but are provided to the decoder unordered. In contrast, our model has visual information as additional input and we expect that the decoder should refer to the image for such kind of information instead of the extracted concepts.

\subsection{Stepwise Image-Topic Merging Decoder}

The essential component of the decoder is the proposed stepwise image-topic merging network. The decoder is based on an LSTM \cite{hochreiter1997long}. At each time step, it combines the textual caption, the attentive visual information, and the attentive conceptual information as the context for generating an output word. The goal is achieved by three modules, the visual attention, the topic attention, and the merging gate.
%
\paragraph{Visual Attention as Output}

The visual attention attends to attracting parts of the image based on the state of the LSTM decoder. In existing work \cite{xu2015show}, only the previous hidden state $\vec{h}_{t-1} \in \R^d$ of the LSTM is used in computation of the visual attention:
\begin{align}
\vec{Z_t} &= \text{tanh} ( \vec{W}^{Z,V} \vec{V}  \oplus \vec{W}^{Z,h} \vec{h}_{t-1}) \\
\vec{\alpha}_t &= \text{softmax}(\vec{Z_t} \vec{w}^{\alpha,Z})
\end{align}
where $\vec{W}^{Z,V} \in \R^{k \times g}, \vec{W}^{Z,h} \in \R^{k\times d}, \vec{w}^{\alpha,Z} \in \R^{k}$ are the learnable parameters. We denote the matrix-vector addition as $\oplus$, which is calculated by adding the vector to each column of the matrix. $\vec{\alpha}_t \in \R^{k}$ is the attentive weights of $\vec{V}$ and the attentive visual input $\vec{z}_t \in \R^{g}$ is calculated as
\begin{align}
\vec{z}_t = \vec{V} \vec{\alpha}_t
\end{align}
The visual input $\vec{z}_t$ and the embedding of the previous output word $\vec{y}_{t-1}$ are the input of the LSTM. 
\begin{align}
\vec{h}_t = \text{LSTM}(\begin{bmatrix}\vec{z}_t\\\vec{y}_{t-1}\end{bmatrix}, \vec{h}_{t-1})
\end{align}
However, there is a noticeable drawback that the previous output word $y_{t-1}$, which is a much stronger indicator than the previous hidden state $\vec{h}_{t-1}$, is not used in the attention. As $\vec{z}_t$ is used as the input, we call it \textbf{input attention}. To overcome that drawback, we add another attention that incorporates the current hidden state $\vec{h}_{t}$, which is based on the last generated word $y_{t-1}$:
\begin{align}
\widetilde{\vec{Z}}_t &= \text{tanh} ( \widetilde{\vec{W}}^{Z,V} \vec{V}  \oplus \widetilde{\vec{W}}^{Z,h} \vec{h}_{t}) \\
\widetilde{\vec{\alpha}}_t &= \text{softmax}(\widetilde{\vec{Z}}_t \widetilde{\vec{w}}^{\alpha,Z}) \\
\widetilde{\vec{z}}_t & = \vec{V} \widetilde{\vec{\alpha}}_t
\end{align}
The procedure resembles the input attention, and we call it \textbf{output attention}. It is worth mentioning that the output attention is essentially the same with the spatial visual attention proposed by \citet{lu2017knowing}. However, they did not see it from the input-output point of view nor combine it with the input attention.

The attentive visual output is further transformed to $\vec{r}_t = \text{tanh}(\vec{W}^{s,z} \widetilde{\vec{z}}_t), \vec{W}^{s,z} \in \R^{e \times g}$, which is of the same dimension as the topic word embedding to simplify the following procedure.

\paragraph{Topic Attention}

In an image caption, different parts concern different topics. In the existing work \cite{you2016image}, the conceptual information is attended based on the previous output word:
\begin{align}
\vec{\beta}_t = \text{softmax} (\vec{T}^\mathsf{T} \vec{U} \vec{y}_{t-1})
\end{align}
where $\vec{U} \in \R^{e \times e}, \vec{\beta}_t \in \R^{m}$. The profound issue is that this approach neglects the visual information. It should be beneficial to provide the attentive visual information when selecting topics. The hidden state of the LSTM contains both the information of previous words and the attentive input visual information. Therefore, the model attends to the topics based on the hidden state of the LSTM:
\begin{align}
\vec{Q}_t &= \text{tanh} (\vec{W}^{Q, T} \vec{T} \oplus \vec{W}^{Q,h} \vec{h}_t)  \label{eq:score1}\\
\vec{\beta}_t &= \text{softmax} (\vec{Q}_t \vec{w}^{\beta,Q}) \label{eq:score2}
\end{align}
where $\vec{W}^{Q,T} \in \R^{m\times e}, \vec{W}^{Q,h} \in \R^{m\times d}, \vec{w}^{\beta, Q} \in \R^{m}$ are the parameters to be learned. $\vec{\beta}_t \in \R^m$ is the weight of the topics, from which the attentive conceptual output $\vec{q}_t \in \R^{e}$ is calculated:
\begin{align}
\vec{q}_t = \vec{T} \vec{\beta}_t
\end{align}
The topic attention $\vec{q}_t$ and the hidden state $\vec{h}_t$ are combined as the contextual information $\vec{s}_t$:
\begin{align}
\vec{s}_t = \text{tanh}(\vec{W}^{s,q} \vec{q}_t + \vec{W}^{s,h} \vec{h}_t)
\end{align}
where $\vec{W}^{s,q} \in \R^{e \times e}, \vec{W}^{s,h} \in \R^{e \times d}$ are learnable parameters.

\paragraph{Merging Gate}

We have prepared both the visual information $\vec{r}_t$ and the contextual information $\vec{s}_t$. It is not reasonable to treat the two kinds of information equally when the decoder generates different types of words. For example, when generating descriptive words (e.g., \textit{behind}, \textit{red}), $\vec{r}_t$ should matter more than $\vec{s}_t$. However, when generating object words (e.g., \textit{people}, \textit{table}), $\vec{s}_t$ is more important. 
We introduce a novel score-based merging mechanism to make the model adaptively learn to adjust the balance:
\begin{align}
\gamma_t &=\sigma(S(\vec{s}_t)-S(\vec{r}_t))\\
\vec{c}_t      &= \gamma_t \vec{s}_t+ (1-\gamma_t) \vec{r}_t
\end{align}
where $\sigma$ is the sigmoid function, $\gamma_t \in [0, 1]$ indicates how important the topic attention is compared to the visual attention, and $S$ is the scoring function. The scoring function needs to evaluate the importance of the topic attention. Noticing that Eq. (\ref{eq:score1}) and Eq. (\ref{eq:score2}) have a similar purpose, we define $S$ similarly:
\begin{align}
S(\vec{s}_t) &=  \text{tanh} (\vec{W}^{S,h} \vec{h}_t + \vec{W}^{S,s} \vec{s}_t ) \cdot \vec{w}^{S} \\
S(\vec{r}_t) &=  \text{tanh} (\vec{W}^{S,h} \vec{h}_t + \vec{W}^{S,r} \vec{r}_t ) \cdot \vec{w}^{S}
\end{align}
where $\cdot$ denotes dot product of vectors, $\vec{W}^{S,s} \in \R^{m \times e}, \vec{W}^{S,r} \in \R^{m \times e}$ are the parameters to be learned, and $\vec{W}^{S,h}, \vec{w}^{s}$ share the weights of $\vec{W}^{Q,h}, \vec{w}^{\beta, Q}$ from Eq. (\ref{eq:score1}) and Eq. (\ref{eq:score2}), respectively. 

Finally, the output word is generated by:
\begin{align}
y_t \sim \vec{p}_t = \text{softmax}(\vec{W}^{p, c} \vec{c}_t)
\end{align}
where each value of $\vec{p}_t \in \R^{|D|}$ is a probability indicating how likely the corresponding word in vocabulary $D$ is the current output word. The whole model is trained using maximum log likelihood and the loss function is the cross entropy loss.

In all, our proposed approach encourages the model to take advantage of all the available information. The adaptive merging mechanism makes the model weigh the information elaborately.

\begin{table*}[t]
\centering
\footnotesize
\begin{tabular}{@{}l c c c c c@{}}
\toprule
Flickr30k                                     & SPICE     & CIDEr     & METEOR    & ROUGE-L   & BLEU-4    \\ \midrule
HardAtt \cite{xu2015show}                     & -         & -         & 0.185     & -         & 0.199     \\
SCA-CNN \cite{chen2017scacnn}                 & -         & -         & 0.195     & -         & 0.223     \\
ATT-FCN \cite{you2016image}                   & -         & -         & 0.189     & -         & 0.230     \\
SCN-LSTM \cite{gan2017semantic}               & -         & -         & 0.210     & -         & 0.257     \\
AdaAtt \cite{lu2017knowing}                   & 0.145     & 0.531     & 0.204     & 0.467     & 0.251     \\
NBT \cite{lu2018neural}                       & 0.156     & 0.575     & 0.217     & -         & 0.271 \\ \midrule
SR-PL \cite{liu2018show}\ssymbol{1}\ssymbol{2} & 0.158     & \bf 0.650     & 0.218     & \bf 0.499     &\bf 0.293     \\ \midrule
simNet                                           & \bf 0.160 & 0.585 & \bf 0.221 &  0.489 & 0.251     \\ \bottomrule
\end{tabular}
\caption{Performance on the Flickr30k Karpathy test split. The symbol \ssymbol{1} denotes directly optimizing CIDEr. The symbol \ssymbol{2} denotes using extra data for training, thus not directly comparable. Nonetheless, our model supersedes all existing models in SPICE, which correlates the best with human judgments.} \label{tab:res-f30k}
\end{table*}

\begin{table*}[t]

\centering
\footnotesize

\begin{tabular}{@{}l c c c c c@{}}
\toprule
COCO                                                 & SPICE     & CIDEr     & METEOR    & ROUGE-L   & BLEU-4    \\ \midrule
HardAtt \cite{xu2015show}                               &-           & -         & 0.230     & -         & 0.250     \\
ATT-FCN \cite{you2016image}                             &-           & -         & 0.243     & -         & 0.304     \\
SCA-CNN \cite{chen2017scacnn}                           & -         & 0.952     & 0.250     & 0.531     & 0.311     \\
LSTM-A \cite{yao2017boosting}                           & 0.186     & 1.002     & 0.254     & 0.540     & 0.326     \\
SCN-LSTM \cite{gan2017semantic}                         & -         & 1.012     & 0.257     & -         & 0.330     \\
Skeleton \cite{wang2017skeleton}                        & -         & 1.069     & 0.268     & 0.552     & 0.336     \\
AdaAtt  \cite{lu2017knowing}                            & 0.195     & 1.085     & 0.266     & 0.549     & 0.332     \\
NBT \cite{lu2018neural}                                 & 0.201     & 1.072     & 0.271     & -         & 0.347     \\ \midrule
DRL \cite{ren2017deep}\ssymbol{1}                       & -         & 0.937     & 0.251     & 0.525     & 0.304     \\
TD-M-ATT \cite{chen2018temporal}\ssymbol{1}             & -         & 1.116     & 0.268     & 0.555     & 0.336     \\
SCST \cite{rennie2017self}\ssymbol{1}                   &-           & 1.140     & 0.267     & 0.557     & 0.342     \\
SR-PL \cite{liu2018show}\ssymbol{1}\ssymbol{2}          & 0.210     & 1.171     & 0.274     & \bf 0.570 & 0.358     \\
Up-Down \cite{anderson2018bottom}\ssymbol{1}\ssymbol{2} & 0.214     & \bf 1.201 & 0.277     & 0.569     & \bf 0.363 \\ \midrule
simNet                                                     & \bf 0.220 & 1.135     & \bf 0.283 & 0.564     & 0.332     \\ \bottomrule
\end{tabular}
\caption{Performance on the COCO Karpathy test split. Symbols, \ssymbol{1} and \ssymbol{2}, are defined similarly. Our model outperforms the current state-of-the-art Up-Down substantially in terms of SPICE.\label{tab:res-mscoco}}
\end{table*}

\section{Experiment}

We describe the datasets and the metrics used for evaluation, followed by the training details and the evaluation of the proposed approach. 

\subsection{Datasets and Metrics}

There are several datasets containing images and their captions. We report results on the popular Microsoft COCO \cite{chen2015microsoft} dataset and the Flickr30k \cite{young2014image} dataset. They contain 123,287 images and 31,000 images, respectively, and each image is annotated with 5 sentences. We report results using the widely-used publicly-available splits in the work of \citet{karpathy2014deep}. There are 5,000 images each in the validation set and the test set for COCO, 1,000 images for Flickr30k.

We report results using the COCO captioning evaluation toolkit \cite{chen2015microsoft} that reports the widely-used automatic evaluation metrics SPICE, CIDEr, BLEU, METEOR, and ROUGE. SPICE \cite{anderson2016spice}, which is based on scene graph matching, and CIDEr \cite{vedantam2015cider}, which is based on n-gram matching, are specifically proposed for evaluating image captioning systems. They both incorporate the consensus of a set of references for an example. BLEU \cite{papineni2002bleu} and METOR \cite{banerjee2005meteor} are originally proposed for machine translation evaluation. ROUGE \cite{lin2003automatic,lin2004rouge} is designed for automatic evaluation of extractive text summarization. In the related studies, it is concluded that SPICE correlates the best with human judgments with a remarkable margin over the other metrics, and is expert in judging detailedness, where the other metrics show negative correlations, surprisingly; CIDEr and METEOR follows with no particular precedence, followed by ROUGE-L, and BLEU-4, in that order \cite{anderson2016spice,vedantam2015cider}.

\subsection{Settings} 

Following common practice, the CNN used is the ResNet152 model \cite{he2016deep} pre-trained on ImageNet.\footnote{We use the pre-trained model from \href{https://github.com/pytorch/vision}{\texttt{torchvision}}.} There are 2048 $7 \times 7$ feature maps, and we project them into 512 feature maps, i.e. $g$ is 512. The word embedding size $e$ is 256 and the hidden size $d$ of the LSTM is 512. We only keep caption words that occur at least 5 times in the training set, resulting in 10,132 words for COCO and 7,544 for Flickr30k. We use the topic extractor pre-trained by \citet{fang2015captions} for 1,000 concepts on COCO. We only use 568 manually-annotated object words as topics. For an image, only the top 5 topics are selected, which means $m$ is 5. The same topic extractor is used for Flickr30k, as COCO provides adequate generality. The caption words and the topic words share the same embeddings. In training, we first train the model without visual attention (freezing the CNN parameters) for 20 epochs with the batch size of 80. The learning rate for the LSTM is 0.0004. Then, we switch to jointly
train the full model with a learning rate of 0.00001, which exponentially decays with the number of epochs so that it is halved every 50 epochs. We also use momentum of 0.8 and weight decay of 0.999. We use Adam \cite{kingma2014adam} for parameter optimization. For fair comparison, we adopt early stop based on CIDEr within maximum 50 epochs.

\begin{table*}[t]
\centering
\scriptsize
\setlength{\tabcolsep}{3pt}
\begin{tabular}{@{}l c c c c c c c c c c c@{}}
	\toprule
	\multirow{2}{*}[-3pt]{Methods}                          &                             \multicolumn{7}{c}{SPICE}                              & \multirow{2}{*}[-3pt]{CIDEr} & \multirow{2}{*}[-3pt]{METEOR} & \multirow{2}{*}[-3pt]{ROUGE-L} & \multirow{2}{*}[-3pt]{BLEU-4} \\
	\cmidrule(lr){2-8}                                      & All       & Objects   & Attributes & Relations & Color     & Count     & Size      &             &               &               &               \\ \midrule
	Baseline (Plain Encoder-Decoder Network)                & 0.150     & 0.295     & 0.048      & 0.039     & 0.022     & 0.004     & 0.023     & 0.762       & 0.220         & 0.495         & 0.251         \\
	Up-Down \cite{anderson2018bottom}\ssymbol{1}\ssymbol{2} & 0.214     & 0.391     & 0.100      & 0.065     & 0.114     & 0.184     & 0.032     & \bf 1.201   & 0.277         & \bf 0.569     & \bf 0.363     \\ \midrule
	Baseline + Input Att.                                   & 0.164     & 0.316     & 0.060      & 0.044     & 0.030     & 0.038     & 0.024     & 0.840       & 0.233         & 0.512         & 0.273         \\
	Baseline + Output Att.                                  & 0.181     & 0.329     & 0.094      & 0.053     & 0.089     & 0.184     & 0.044     & 0.968       & 0.253         & 0.534         & 0.301         \\
	Baseline + Input Att.  + Output Att.                    & 0.187     & 0.338     & 0.101      & 0.055     & \bf 0.115 & 0.161     & \bf 0.048 & 1.038       & 0.259         & 0.542         & 0.311         \\ \midrule
	Baseline + Topic Att.                                   & 0.184     & 0.348     & 0.074      & 0.051     & 0.047     & 0.064     & 0.037     & 0.915       & 0.250         & 0.517         & 0.260         \\
	Baseline + Topic Att. + MGate                           & 0.189     & 0.355     & 0.080      & 0.051     & 0.055     & 0.090     & 0.033     & 0.959       & 0.256         & 0.527         & 0.281         \\ \midrule
	Baseline + Input Att. + Output Att. + Topic Att.        & 0.206     & 0.381     & 0.091      & 0.060     & 0.075     & 0.094     & 0.045     & 1.068       & 0.273         & 0.556         & 0.320         \\ \midrule
	simNet (Full Model)                                     & \bf 0.220 & \bf 0.394 & \bf 0.109  & \bf 0.070 & 0.088     & \bf 0.202 & 0.045     & 1.135       & \bf 0.283     & 0.564         & 0.332         \\ \bottomrule
\end{tabular}
\caption{Results of incremental analysis. For a better understanding of the differences, we further list the breakdown of SPICE F-scores. \textit{Objects} indicates comprehensiveness, and the others indicate detailedness. Additionally, we report the performance of the current state-of-the-art Up-Down for further comparison, which uses extra dense-annotated data for pre-training and directly optimizes CIDEr.}
\label{tab:res-ab}
\end{table*}

\subsection{Results} 

We compare our approach with various representative systems on Flickr30k and COCO, including the recently proposed NBT that is the state-of-the-art on the two datasets in comparable settings. Table \ref{tab:res-f30k} shows the result on Flickr30k. As we can see, our model outperforms the comparable systems in terms of all of the metrics except BLEU-4. Moreover, our model overpasses the state-of-the-art with a comfortable margin in terms of SPICE, which is shown to correlate the best with human judgments \cite{anderson2016spice}. 

Table \ref{tab:res-mscoco} shows the results on COCO. Among the directly comparable models, our model is arguably the best and outperforms the existing models except in terms of BLEU-4. Most encouragingly, our model is also competitive with Up-Down \cite{anderson2018bottom}, which uses much larger dataset, Visual Genome \cite{krishna2017visual}, with dense annotations to train the object detector, and directly optimizes CIDEr. Especially, our model outperforms the state-of-the-art substantially in SPICE and METEOR. Breakdown of SPICE F-scores over various subcategories (see Table~\ref{tab:res-ab}) shows that our model is in dominant lead in almost all subcategories. It proves the effectiveness of our approach and indicates that our model is quite data efficient. 

For the methods that directly optimize CIDEr, it is intuitive that CIDEr can improve significantly. The similar improvement of BLEU-4 is evidence that optimizing CIDEr leads to more n-gram matching. However, it comes to our notice that the improvements of SPICE, METEOR, and ROUGE-L are far less significant, which suggests there may be a gaming situation where the n-gram matching is wrongfully exploited by the model in reinforcement learning. As shown by \citet{liu2017improved}, it is most reasonable to jointly optimize all the metrics at the same time.

We also evaluate the proposed model on the COCO evaluation server, the results of which are shown in Appendix \ref{sec:coco-server}, due to limited space.

\begin{table}[t]
\centering
\footnotesize
\begin{tabular}{@{}l c c c@{}}
\toprule
Method             & Precision & Recall          & F1    \\ \midrule
Topics ($m$=5)      & 49.95     & 38.91          & 42.48 \\ \midrule
All words ($m$=5)  & \bf84.01     & 17.99 & 29.49\\
All words ($m$=10) & 70.90     & 30.18 & 42.05 \\
All words ($m$=20) &  52.51     & \bf 44.53           & \bf 47.80 \\ \bottomrule
\end{tabular}
\caption{Performance of visual word extraction.}
\label{tab:res-topic-extract}
\end{table}

\begin{table}[t]
\centering
\footnotesize
\setlength{\tabcolsep}{3pt}
\begin{tabular}{@{}l c c c c c@{}}
\toprule
Method             & S     & C     & M     & R     & B     \\ \midrule
Topics ($m$=5)      & \bf 0.220 & \bf 1.135 & \bf 0.283 & \bf 0.564 & \bf 0.332 \\ \midrule
All words ($m$=5)  & 0.197 & 1.047 & 0.264 & 0.550 & 0.314 \\
All words ($m$=10) & 0.201 & 1.076 & 0.256 & 0.528 & 0.293 \\
All words ($m$=20) & 0.209 & 1.117 & 0.276 & 0.561 & 0.329 \\ \bottomrule
\end{tabular}
\caption{Effect of using different visual words.}
\label{tab:res-topic}
\end{table}

\section{Analysis}

In this section, we analyze the contribution of each component in the proposed approach, and give examples to show the strength and the potential improvements of the model. The analysis is conducted on the test set of COCO.

\paragraph{Topic Extraction}

The motivation of using objects as topics is that they are easier to identify so that the generation suffers less from erroneous predictions. This can be proved by the F-score of the identified topics in the test set, which is shown in Table \ref{tab:res-topic-extract}. Using top-5 object words is at least as good as using top-10 all words. However, using top-10 all words introduces more erroneous visual words to the generation. As shown in Table \ref{tab:res-topic}, when extracting all words, providing more words to the model indeed increases the captioning performance. However, even when top-20 all words are used, the performance is still far behind using only top-5 object words and seems to reach the performance ceiling. It proves that for semantic attention, it is also important to limit the absolute number of incorrect visual words instead of merely the precision or the recall. It is also interesting to check whether using other kind of words can reach the same effect. Unfortunately, in our experiments, only using verbs or adjectives as semantic concepts works poorly.

\begin{figure}[t]

\centering
\includegraphics[width=0.35\textwidth]{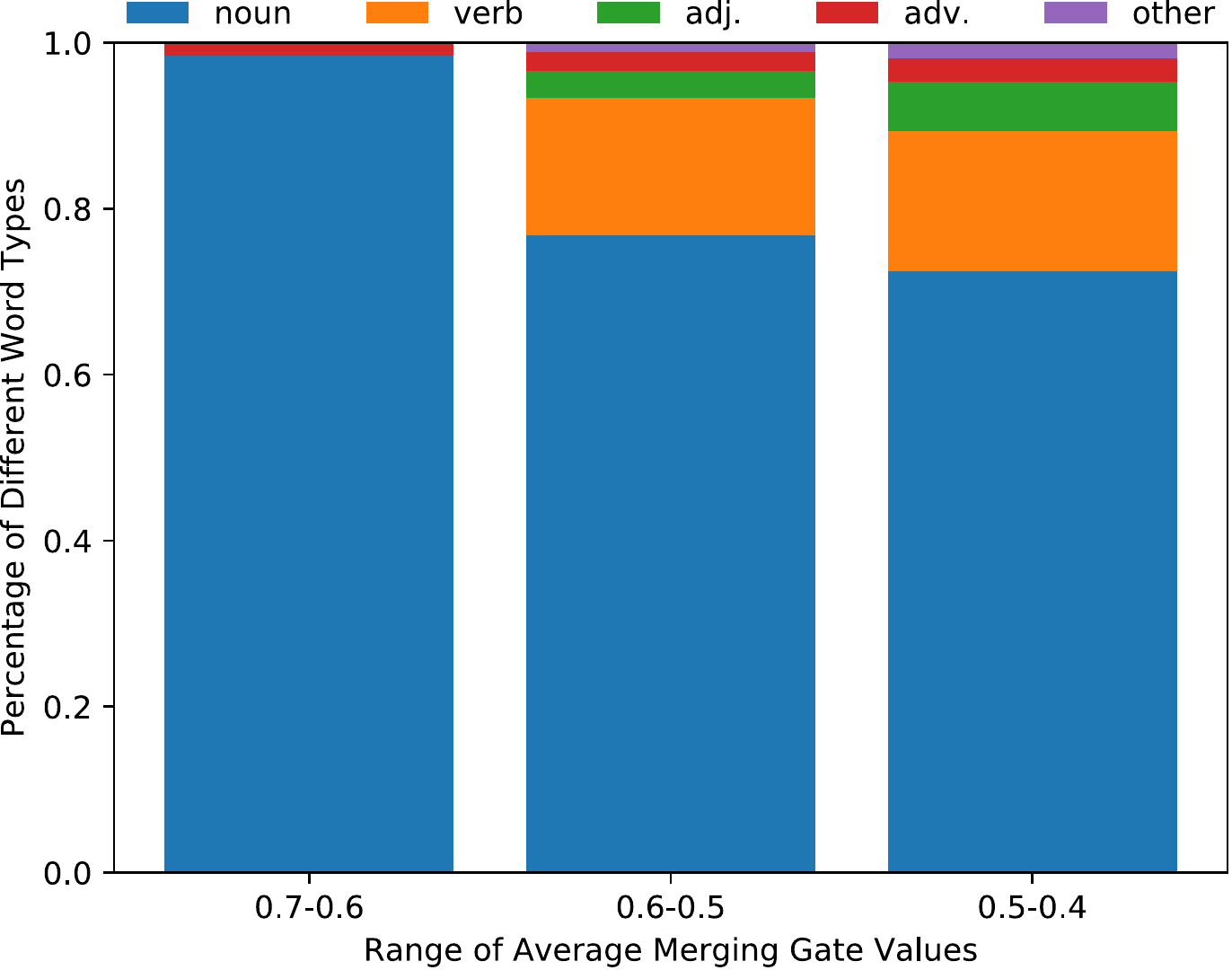}
\caption{Average merging gate values according to word types. As we can see, object words (noun) dominate the high value range, while attribute and relation words are assigned lower values, indicating the merging gate learns to efficiently combine the information.}
\label{fig:bgate}
\end{figure}

\begin{figure*}[t]

\centering
\includegraphics[width=1.0\textwidth]{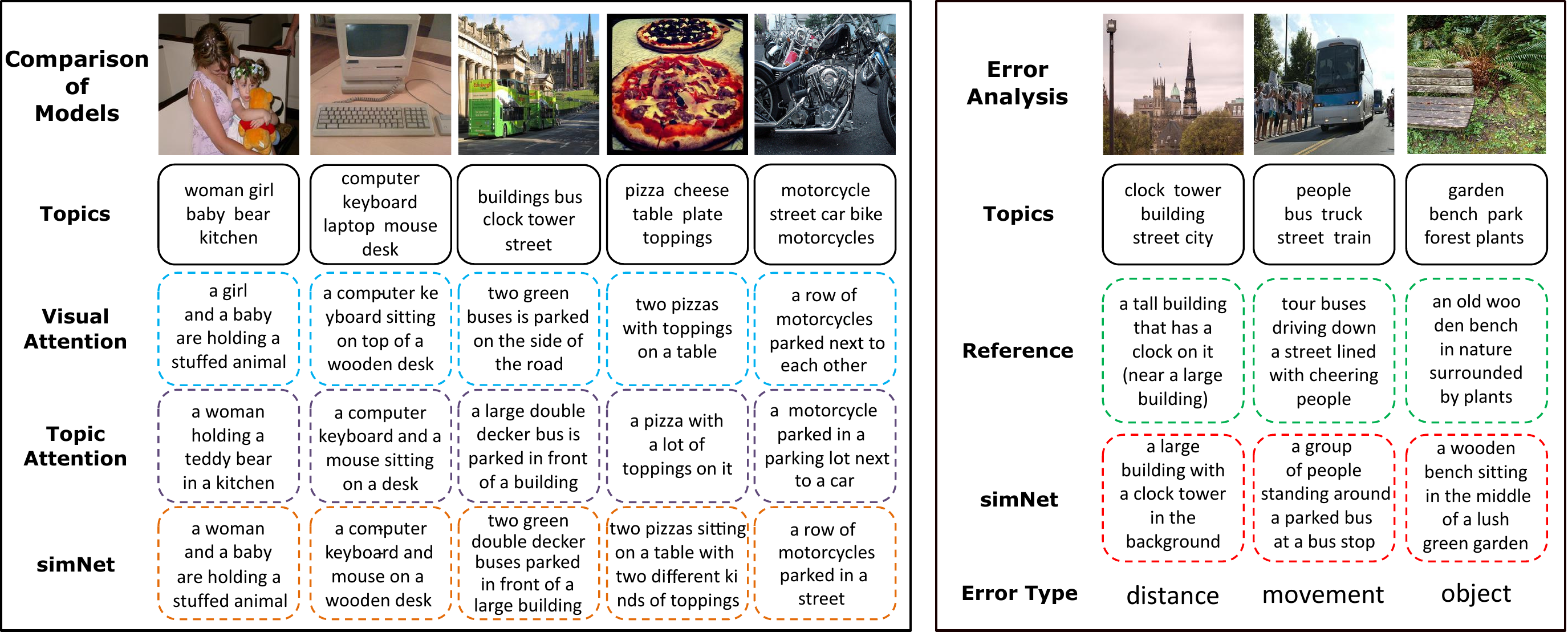}

\caption{Examples of the generated captions. The left plot compares simNet with visual attention and topic attention. Visual attention is good at portraying the relations but is less specific in objects. Topic attention includes more objects but lacks details, such as material, color, and number. The proposed model achieves a very good balance. The right plot shows the error analysis of the proposed simNet.}
\label{fig:ex-comp}
\end{figure*}

To examine the contributions of the sub-modules in our model, we conduct a series of experiments. The results are summarized in Table \ref{tab:res-ab}. To help with the understanding of the differences, we also report the breakdown of SPICE F-scores.

\paragraph{Visual Attention} Our input attention achieves similar results to previous work \cite{xu2015show}, if not better. Using only the output attention is much more effective than using only the input attention, with substantial improvements in all metrics, showing the impact of information gap caused by delayed input in attention. Combining the input attention and the output attention can further improve the results, especially in color and size descriptions.

\paragraph{Topic Attention} As expected, compared with visual attention, the topic attention is better at identifying objects but worse at identifying attributes. We also apply the merging gate to the topic attention, but it now merges $\vec{q_t}$ and $\vec{h_t}$ instead of $\vec{s_t}$ and $\vec{r_t}$. With the merging gate, the model can balance the information in caption text and extracted topics, resulting in better overall scores. While it overpasses the conventional visual attention, it lags behind the output attention.

\paragraph{Merging Gate} Combing the visual attention and the topic attention directly indeed results in a huge boost in performance, which confirms our motivation. However, directly combining them also causes lower scores in attributes, color, count, and size, showing that the advantages are not fully made use of. The most dramatic improvements come from applying the merging gate to the combined attention, showing that the proposed balance mechanism can adaptively combine the two kinds of information and is essential to the overall performance. The average merging gate value summarized in Figure \ref{fig:bgate} suggests the same. 

\medskip

We give some examples in the left plot of Figure~\ref{fig:ex-comp} to illustrate the differences between the models more intuitively. From the examples, it is clear that the proposed simNet generates the best captions in that more objects are described and many informative and detailed attributes are included, such as the quantity and the color.

\begin{figure*}[ht]

\centering
\includegraphics[width=1.0\textwidth]{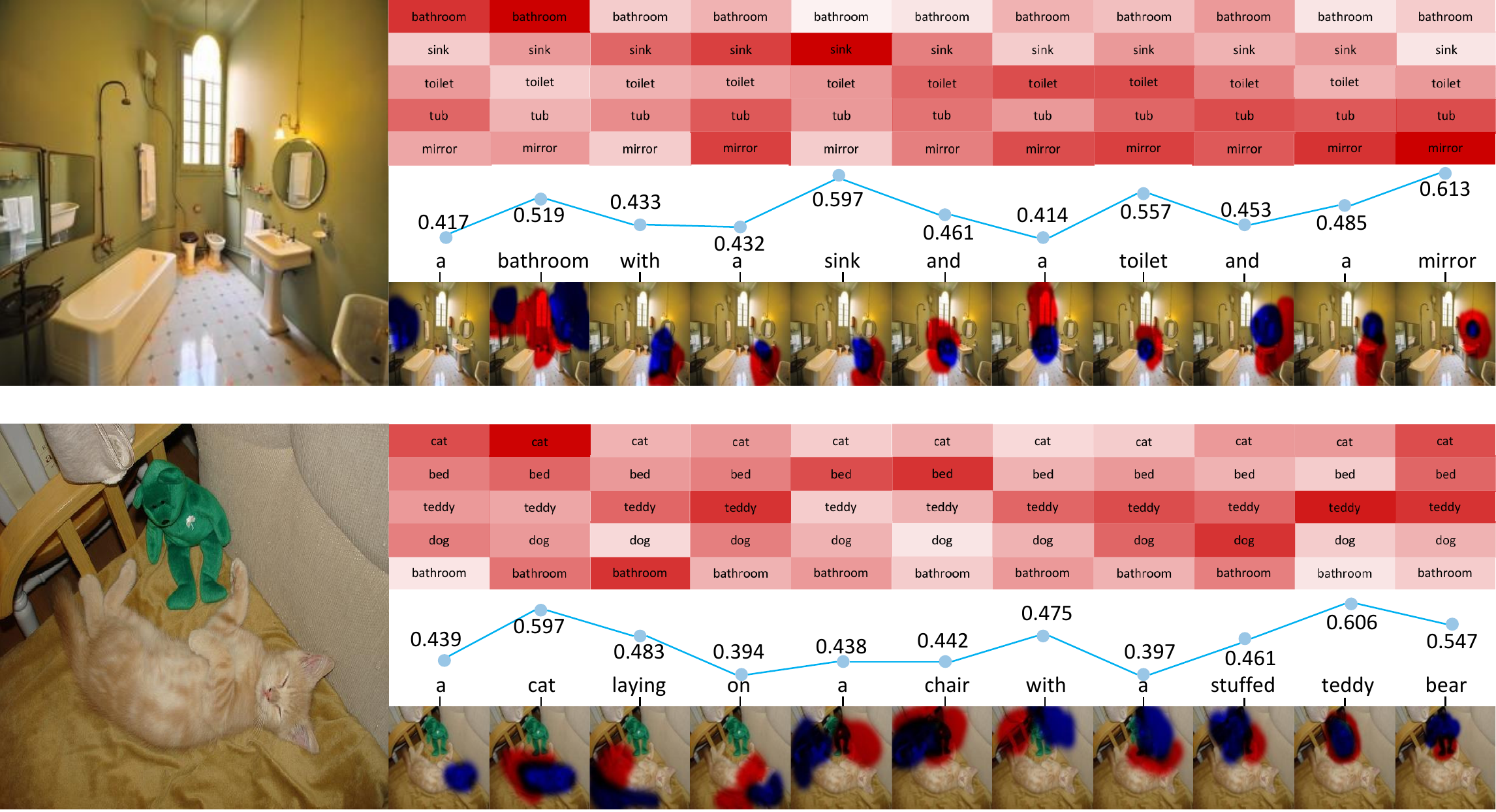}

\caption{Visualization. Please view in color. Here, we give two running examples. The upper part of each example shows the attention weights of each of 5 extracted topics. Deeper color means larger in value. The middle part shows the value of the merging gate that determines the importance of the topic attention. The lower part shows the visualization of visual attention. The attended region is covered with color. The blue shade indicates the output attention. The red shade indicates the input attention. }
\label{fig:ex-vis}
\end{figure*}

\paragraph{Visualization}

Figure \ref{fig:ex-vis} shows the visualization of the topic attention and the visual attention with running examples. As we can see, the topic attention is active when generating a phrase containing the related topic. For example, \textit{bathroom} is always most attended when generating \textit{a bathroom}. The merging gate learns to direct the information flow efficiently. When generating words such as \textit{on} and \textit{a}, it gives lower weight to the topic attention and prefers the visual attention. As to the visual attention, the output attention is much more focused than the input attention. As we hypothesized, the conventional input attention lacks the information of the last generated word and does not know what to look for exactly. For example, when generating \textit{bathroom}, the input attention does not know the previous generated word is \textit{a}, and it loses its focus, while the output attention is relatively more concentrated. Moreover, the merging gate learns to overcome the erroneous topics, as shown in the second example. When generating \textit{chair}, the topic attention is focused on a wrong object \textit{bed}, while the visual attention attends correctly to the chair, and especially the output attention attends to the armrest. The merging gate effectively remedies the misleading information from the topic attention and outputs a lower weight, resulting in the model correctly generating the word \textit{chair}. 



\paragraph{Error Analysis}
We conduct error analysis using the proposed (full) model on the test set to provide insights on how the model may be improved. We find 123 out of 1000 generated captions that are not satisfactory. There are mainly three types of errors, i.e. distance (32, 26\%), movement (22, 18\%), and object (60, 49\%), with 9 (7\%) other errors. Distance error takes place when there is a lot of objects and the model cannot grasp the foreground and the background relationship. Movement error means that the model fails to describe whether the objects are moving. Those two kinds of errors are hard to eliminate, as they are fundamental problems of computer vision waiting to be resolved. Object error happens when there are incorrect extracted topics, and the merging gate regards the topic as grounded in the image. In the given example, the incorrect topic is \textit{garden}. The tricky part is that the topic is seemingly correct according to the image features or otherwise the proposed model will choose other topics. A more powerful topic extractor may help with the problem but it is unlikely to be completely avoided.

\section{Conclusions}

We propose the stepwise image-topic merging network to sequentially and adaptively merge the visual and the conceptual information for improved image captioning. To our knowledge, we are the first to combine the visual and the semantic attention to achieve substantial improvements. We introduce the stepwise merging mechanism to efficiently guide the two kinds of information when generating the caption. The experimental results demonstrate the effectiveness of the proposed approach, which substantially outperforms the state-of-the-art image captioning methods in terms of SPICE on COCO and Flickr30k datasets. Quantitative and qualitative analysis show that the generated captions are both detailed and comprehensive in comparison with the existing methods.

\section*{Acknowledgments}

This work was supported in part by National Natural Science Foundation of China (No. 61673028). We thank all the anonymous reviewers for their constructive comments and suggestions. Xu Sun is the corresponding author of this paper.

\bibliography{emnlp2018}

\begin{thebibliography}{48}
\expandafter\ifx\csname natexlab\endcsname\relax\def\natexlab#1{#1}\fi

\bibitem[{Anderson et~al.(2016)Anderson, Fernando, Johnson, and
  Gould}]{anderson2016spice}
Peter Anderson, Basura Fernando, Mark Johnson, and Stephen Gould. 2016.
\newblock {SPICE:} semantic propositional image caption evaluation.
\newblock In \emph{Computer Vision - {ECCV} 2016 - 14th European Conference,
  Amsterdam, The Netherlands, October 11-14, 2016, Proceedings, Part {V}},
  volume 9909 of \emph{Lecture Notes in Computer Science}, pages 382--398.
  Springer.

\bibitem[{Anderson et~al.(2018)Anderson, He, Buehler, Teney, Johnson, Gould,
  and Zhang}]{anderson2018bottom}
Peter Anderson, Xiaodong He, Chris Buehler, Damien Teney, Mark Johnson, Stephen
  Gould, and Lei Zhang. 2018.
\newblock Bottom-up and top-down attention for image captioning and {VQA}.
\newblock In \emph{2018 {IEEE} Conference on Computer Vision and Pattern
  Recognition, {CVPR} 2018, Salt Lake City, UT, USA, June 18-22, 2018}.

\bibitem[{Bahdanau et~al.(2014)Bahdanau, Cho, and Bengio}]{bahdanau2014neural}
Dzmitry Bahdanau, Kyunghyun Cho, and Yoshua Bengio. 2014.
\newblock Neural machine translation by jointly learning to align and
  translate.
\newblock \emph{CoRR}, abs/1409.0473.

\bibitem[{Banerjee and Lavie(2005)}]{banerjee2005meteor}
Satanjeev Banerjee and Alon Lavie. 2005.
\newblock {METEOR:} an automatic metric for {MT} evaluation with improved
  correlation with human judgments.
\newblock In \emph{Proceedings of the Workshop on Intrinsic and Extrinsic
  Evaluation Measures for Machine Translation and/or Summarization@ACL 2005,
  Ann Arbor, Michigan, June 29, 2005}, pages 65--72. Association for
  Computational Linguistics.

\bibitem[{Chen et~al.(2018)Chen, Ding, Zhao, and Han}]{chen2018temporal}
Hui Chen, Guiguang Ding, Sicheng Zhao, and Jungong Han. 2018.
\newblock Temporal-difference learning with sampling baseline for image
  captioning.
\newblock In \emph{Proceedings of the Thirty-Second {AAAI} Conference on
  Artificial Intelligence, New Orleans, Louisiana, USA, February 2-7, 2018}.
  {AAAI} Press.

\bibitem[{Chen et~al.(2017)Chen, Zhang, Xiao, Nie, Shao, Liu, and
  Chua}]{chen2017scacnn}
Long Chen, Hanwang Zhang, Jun Xiao, Liqiang Nie, Jian Shao, Wei Liu, and
  Tat{-}Seng Chua. 2017.
\newblock {SCA-CNN:} spatial and channel-wise attention in convolutional
  networks for image captioning.
\newblock In \emph{2017 {IEEE} Conference on Computer Vision and Pattern
  Recognition, {CVPR} 2017, Honolulu, HI, USA, July 21-26, 2017}, pages
  6298--6306. {IEEE} Computer Society.

\bibitem[{Chen et~al.(2015)Chen, Fang, Lin, Vedantam, Gupta, Doll{\'{a}}r, and
  Zitnick}]{chen2015microsoft}
Xinlei Chen, Hao Fang, Tsung{-}Yi Lin, Ramakrishna Vedantam, Saurabh Gupta,
  Piotr Doll{\'{a}}r, and C.~Lawrence Zitnick. 2015.
\newblock Microsoft {COCO} captions: Data collection and evaluation server.
\newblock \emph{CoRR}, abs/1504.00325.

\bibitem[{Chen and Zitnick(2015)}]{chen2015mind}
Xinlei Chen and C.~Lawrence Zitnick. 2015.
\newblock Mind's eye: {A} recurrent visual representation for image caption
  generation.
\newblock In \emph{2015 {IEEE} Conference on Computer Vision and Pattern
  Recognition, {CVPR} 2015, Boston, MA, USA, June 7-12, 2015}, pages
  2422--2431. {IEEE} Computer Society.

\bibitem[{Das et~al.(2017)Das, Kottur, Gupta, Singh, Yadav, Moura, Parikh, and
  Batra}]{das2017visual}
Abhishek Das, Satwik Kottur, Khushi Gupta, Avi Singh, Deshraj Yadav, Jos{\'{e}}
  M.~F. Moura, Devi Parikh, and Dhruv Batra. 2017.
\newblock Visual dialog.
\newblock In \emph{2017 {IEEE} Conference on Computer Vision and Pattern
  Recognition, {CVPR} 2017, Honolulu, HI, USA, July 21-26, 2017}, pages
  1080--1089. {IEEE} Computer Society.

\bibitem[{Fang et~al.(2015)Fang, Gupta, Iandola, Srivastava, Deng,
  Doll{\'{a}}r, Gao, He, Mitchell, Platt, Zitnick, and
  Zweig}]{fang2015captions}
Hao Fang, Saurabh Gupta, Forrest~N. Iandola, Rupesh~Kumar Srivastava, Li~Deng,
  Piotr Doll{\'{a}}r, Jianfeng Gao, Xiaodong He, Margaret Mitchell, John~C.
  Platt, C.~Lawrence Zitnick, and Geoffrey Zweig. 2015.
\newblock From captions to visual concepts and back.
\newblock In \emph{{IEEE} Conference on Computer Vision and Pattern
  Recognition, {CVPR} 2015, Boston, MA, USA, June 7-12, 2015}, pages
  1473--1482. {IEEE} Computer Society.

\bibitem[{Farhadi et~al.(2010)Farhadi, Hejrati, Sadeghi, Young, Rashtchian,
  Hockenmaier, and Forsyth}]{farhadi2010every}
Ali Farhadi, Seyyed Mohammad~Mohsen Hejrati, Mohammad~Amin Sadeghi, Peter
  Young, Cyrus Rashtchian, Julia Hockenmaier, and David~A. Forsyth. 2010.
\newblock Every picture tells a story: Generating sentences from images.
\newblock In \emph{Computer Vision - {ECCV} 2010, 11th European Conference on
  Computer Vision, Heraklion, Crete, Greece, September 5-11, 2010, Proceedings,
  Part {IV}}, volume 6314 of \emph{Lecture Notes in Computer Science}, pages
  15--29. Springer.

\bibitem[{Gan et~al.(2017)Gan, Gan, He, Pu, Tran, Gao, Carin, and
  Deng}]{gan2017semantic}
Zhe Gan, Chuang Gan, Xiaodong He, Yunchen Pu, Kenneth Tran, Jianfeng Gao,
  Lawrence Carin, and Li~Deng. 2017.
\newblock Semantic compositional networks for visual captioning.
\newblock In \emph{2017 {IEEE} Conference on Computer Vision and Pattern
  Recognition, {CVPR} 2017, Honolulu, HI, USA, July 21-26, 2017}, pages
  1141--1150. {IEEE} Computer Society.

\bibitem[{He et~al.(2016)He, Zhang, Ren, and Sun}]{he2016deep}
Kaiming He, Xiangyu Zhang, Shaoqing Ren, and Jian Sun. 2016.
\newblock Deep residual learning for image recognition.
\newblock In \emph{2016 {IEEE} Conference on Computer Vision and Pattern
  Recognition, {CVPR} 2016, Las Vegas, NV, USA, June 27-30, 2016}, pages
  770--778. {IEEE} Computer Society.

\bibitem[{Hochreiter and Schmidhuber(1997)}]{hochreiter1997long}
Sepp Hochreiter and J{\"{u}}rgen Schmidhuber. 1997.
\newblock Long short-term memory.
\newblock \emph{Neural Computation}, 9(8):1735--1780.

\bibitem[{Karpathy and Li(2015)}]{karpathy2014deep}
Andrej Karpathy and Fei{-}Fei Li. 2015.
\newblock Deep visual-semantic alignments for generating image descriptions.
\newblock In \emph{{IEEE} Conference on Computer Vision and Pattern
  Recognition, {CVPR} 2015, Boston, MA, USA, June 7-12, 2015}, pages
  3128--3137. {IEEE} Computer Society.

\bibitem[{Karpathy and Li(2017)}]{karpathy2017deep}
Andrej Karpathy and Fei{-}Fei Li. 2017.
\newblock Deep visual-semantic alignments for generating image descriptions.
\newblock \emph{{IEEE} Transactions on Pattern Analysis and Machine
  Intelligence}, 39(4):664--676.

\bibitem[{Kingma and Ba(2014)}]{kingma2014adam}
Diederik~P. Kingma and Jimmy Ba. 2014.
\newblock Adam: {A} method for stochastic optimization.
\newblock \emph{CoRR}, abs/1412.6980.

\bibitem[{Krishna et~al.(2017)Krishna, Zhu, Groth, Johnson, Hata, Kravitz,
  Chen, Kalantidis, Li, Shamma, Bernstein, and Li}]{krishna2017visual}
Ranjay Krishna, Yuke Zhu, Oliver Groth, Justin Johnson, Kenji Hata, Joshua
  Kravitz, Stephanie Chen, Yannis Kalantidis, Li{-}Jia Li, David~A. Shamma,
  Michael~S. Bernstein, and Fei{-}Fei Li. 2017.
\newblock Visual genome: Connecting language and vision using crowdsourced
  dense image annotations.
\newblock \emph{International Journal of Computer Vision}, 123(1):32--73.

\bibitem[{Kulkarni et~al.(2013)Kulkarni, Premraj, Ordonez, Dhar, Li, Choi,
  Berg, and Berg}]{kulkarni2013babytalk}
Girish Kulkarni, Visruth Premraj, Vicente Ordonez, Sagnik Dhar, Siming Li,
  Yejin Choi, Alexander~C. Berg, and Tamara~L. Berg. 2013.
\newblock Baby{T}alk: Understanding and generating simple image descriptions.
\newblock \emph{{IEEE} Transactions on Pattern Analysis Machine Intelligence},
  35(12):2891--2903.

\bibitem[{Lin(2004)}]{lin2004rouge}
Chin-Yew Lin. 2004.
\newblock {ROUGE:} a package for automatic evaluation of summaries.
\newblock In \emph{Text Summarization Branches Out: Proceedings of the ACL-04
  Workshop, Barcelona, Spain, July, 2004}, pages 74--81. Association for
  Computational Linguistics.

\bibitem[{Lin and Hovy(2003)}]{lin2003automatic}
Chin{-}Yew Lin and Eduard~H. Hovy. 2003.
\newblock Automatic evaluation of summaries using n-gram co-occurrence
  statistics.
\newblock In \emph{Human Language Technology Conference of the North American
  Chapter of the Association for Computational Linguistics, {HLT-NAACL} 2003,
  Edmonton, Canada, May 27 - June 1, 2003}. The Association for Computational
  Linguistics.

\bibitem[{Lin et~al.(2018)Lin, Sun, Ren, Ma, Su, and Su}]{lin18deconvolution}
Junyang Lin, Xu~Sun, Xuancheng Ren, Shuming Ma, Jinsong Su, and Qi~Su. 2018.
\newblock Deconvolution-based global decoding for neural machine translation.
\newblock In \emph{Proceedings of the 27th International Conference on
  Computational Linguistics, {COLING} 2018, Santa Fe, New Mexico, USA, August
  20-26, 2018}, pages 3260--3271. Association for Computational Linguistics.

\bibitem[{Liu et~al.(2017)Liu, Zhu, Ye, Guadarrama, and
  Murphy}]{liu2017improved}
Siqi Liu, Zhenhai Zhu, Ning Ye, Sergio Guadarrama, and Kevin Murphy. 2017.
\newblock Improved image captioning via policy gradient optimization of spider.
\newblock In \emph{{IEEE} International Conference on Computer Vision, {ICCV}
  2017, Venice, Italy, October 22-29, 2017}, pages 873--881. {IEEE} Computer
  Society.

\bibitem[{Liu et~al.(2018)Liu, Li, Shao, Chen, and Wang}]{liu2018show}
Xihui Liu, Hongsheng Li, Jing Shao, Dapeng Chen, and Xiaogang Wang. 2018.
\newblock Show, tell and discriminate: Image captioning by self-retrieval with
  partially labeled data.
\newblock \emph{CoRR}, abs/1803.08314.

\bibitem[{Lu et~al.(2017)Lu, Xiong, Parikh, and Socher}]{lu2017knowing}
Jiasen Lu, Caiming Xiong, Devi Parikh, and Richard Socher. 2017.
\newblock Knowing when to look: Adaptive attention via a visual sentinel for
  image captioning.
\newblock In \emph{2017 {IEEE} Conference on Computer Vision and Pattern
  Recognition, {CVPR} 2017, Honolulu, HI, USA, July 21-26, 2017}, pages
  3242--3250. {IEEE} Computer Society.

\bibitem[{Lu et~al.(2018)Lu, Yang, Batra, and Parikh}]{lu2018neural}
Jiasen Lu, Jianwei Yang, Dhruv Batra, and Devi Parikh. 2018.
\newblock Neural baby talk.
\newblock In \emph{2018 {IEEE} Conference on Computer Vision and Pattern
  Recognition, {CVPR} 2018, Salt Lake City, UT, USA, June 18-22, 2018}.

\bibitem[{Ma et~al.(2018)Ma, Sun, Lin, and Ren}]{shuming2018}
Shuming Ma, Xu~Sun, Junyang Lin, and Xuancheng Ren. 2018.
\newblock A hierarchical end-to-end model for jointly improving text
  summarization and sentiment classification.
\newblock In \emph{Proceedings of the Twenty-Seventh International Joint
  Conference on Artificial Intelligence, {IJCAI} 2018, July 13-19, 2018,
  Stockholm, Sweden.}, pages 4251--4257. ijcai.org.

\bibitem[{Mao et~al.(2014)Mao, Xu, Yang, Wang, and Yuille}]{mao2014deep}
Junhua Mao, Wei Xu, Yi~Yang, Jiang Wang, and Alan~L. Yuille. 2014.
\newblock Deep captioning with multimodal recurrent neural networks {(m-RNN)}.
\newblock \emph{CoRR}, abs/1412.6632.

\bibitem[{Papineni et~al.(2002)Papineni, Roukos, Ward, and
  Zhu}]{papineni2002bleu}
Kishore Papineni, Salim Roukos, Todd Ward, and Wei{-}Jing Zhu. 2002.
\newblock {BLEU}: a method for automatic evaluation of machine translation.
\newblock In \emph{Proceedings of the 40th Annual Meeting of the Association
  for Computational Linguistics, July 6-12, 2002, Philadelphia, PA, {USA.}},
  pages 311--318. {ACL}.

\bibitem[{Ren et~al.(2015)Ren, He, Girshick, and Sun}]{ren2015faster}
Shaoqing Ren, Kaiming He, Ross~B. Girshick, and Jian Sun. 2015.
\newblock Faster {R-CNN:} towards real-time object detection with region
  proposal networks.
\newblock In \emph{Advances in Neural Information Processing Systems 28: Annual
  Conference on Neural Information Processing Systems 2015, December 7-12,
  2015, Montreal, Quebec, Canada}, pages 91--99.

\bibitem[{Ren et~al.(2017{\natexlab{a}})Ren, He, Girshick, and
  Sun}]{ren2017faster}
Shaoqing Ren, Kaiming He, Ross~B. Girshick, and Jian Sun. 2017{\natexlab{a}}.
\newblock Faster {R-CNN:} towards real-time object detection with region
  proposal networks.
\newblock \emph{{IEEE} Transactions on Pattern Analysis and Machine
  Intelligence}, 39(6):1137--1149.

\bibitem[{Ren et~al.(2017{\natexlab{b}})Ren, Wang, Zhang, Lv, and
  Li}]{ren2017deep}
Zhou Ren, Xiaoyu Wang, Ning Zhang, Xutao Lv, and Li{-}Jia Li.
  2017{\natexlab{b}}.
\newblock Deep reinforcement learning-based image captioning with embedding
  reward.
\newblock In \emph{2017 {IEEE} Conference on Computer Vision and Pattern
  Recognition, {CVPR} 2017, Honolulu, HI, USA, July 21-26, 2017}, pages
  1151--1159. {IEEE} Computer Society.

\bibitem[{Rennie et~al.(2017)Rennie, Marcheret, Mroueh, Ross, and
  Goel}]{rennie2017self}
Steven~J. Rennie, Etienne Marcheret, Youssef Mroueh, Jarret Ross, and Vaibhava
  Goel. 2017.
\newblock Self-critical sequence training for image captioning.
\newblock In \emph{2017 {IEEE} Conference on Computer Vision and Pattern
  Recognition, {CVPR} 2017, Honolulu, HI, USA, July 21-26, 2017}, pages
  1179--1195. {IEEE} Computer Society.

\bibitem[{Vedantam et~al.(2015)Vedantam, Zitnick, and
  Parikh}]{vedantam2015cider}
Ramakrishna Vedantam, C.~Lawrence Zitnick, and Devi Parikh. 2015.
\newblock {CIDEr:} consensus-based image description evaluation.
\newblock In \emph{{IEEE} Conference on Computer Vision and Pattern
  Recognition, {CVPR} 2015, Boston, MA, USA, June 7-12, 2015}, pages
  4566--4575. {IEEE} Computer Society.

\bibitem[{Venugopalan et~al.(2015)Venugopalan, Rohrbach, Donahue, Mooney,
  Darrell, and Saenko}]{venugopalan2015sequence}
Subhashini Venugopalan, Marcus Rohrbach, Jeffrey Donahue, Raymond~J. Mooney,
  Trevor Darrell, and Kate Saenko. 2015.
\newblock Sequence to sequence - video to text.
\newblock In \emph{2015 {IEEE} International Conference on Computer Vision,
  {ICCV} 2015, Santiago, Chile, December 7-13, 2015}, pages 4534--4542. {IEEE}
  Computer Society.

\bibitem[{Vinyals et~al.(2015)Vinyals, Toshev, Bengio, and
  Erhan}]{vinyals2015show}
Oriol Vinyals, Alexander Toshev, Samy Bengio, and Dumitru Erhan. 2015.
\newblock Show and tell: {A} neural image caption generator.
\newblock In \emph{2015 {IEEE} Conference on Computer Vision and Pattern
  Recognition, {CVPR} 2015, Boston, MA, USA, June 7-12, 2015}, pages
  3156--3164. {IEEE} Computer Society.

\bibitem[{Wang et~al.(2017)Wang, Lin, Shen, Cohen, and
  Cottrell}]{wang2017skeleton}
Yufei Wang, Zhe Lin, Xiaohui Shen, Scott Cohen, and Garrison~W. Cottrell. 2017.
\newblock Skeleton key: Image captioning by skeleton-attribute decomposition.
\newblock In \emph{2017 {IEEE} Conference on Computer Vision and Pattern
  Recognition, {CVPR} 2017, Honolulu, HI, USA, July 21-26, 2017}, pages
  7378--7387. {IEEE} Computer Society.

\bibitem[{Wu et~al.(2016)Wu, Shen, Liu, Dick, and van~den Hengel}]{wu2016what}
Qi~Wu, Chunhua Shen, Lingqiao Liu, Anthony~R. Dick, and Anton van~den Hengel.
  2016.
\newblock What value do explicit high level concepts have in vision to language
  problems?
\newblock In \emph{2016 {IEEE} Conference on Computer Vision and Pattern
  Recognition, {CVPR} 2016, Las Vegas, NV, USA, June 27-30, 2016}, pages
  203--212. {IEEE} Computer Society.

\bibitem[{Wu et~al.(2017)Wu, Wieland, Farivar, and Schiller}]{wu2017automatic}
Shaomei Wu, Jeffrey Wieland, Omid Farivar, and Julie Schiller. 2017.
\newblock Automatic alt-text: Computer-generated image descriptions for blind
  users on a social network service.
\newblock In \emph{Proceedings of the 2017 {ACM} Conference on Computer
  Supported Cooperative Work and Social Computing, {CSCW} 2017, Portland, OR,
  USA, February 25 - March 1, 2017}, pages 1180--1192. {ACM}.

\bibitem[{Xu et~al.(2018{\natexlab{a}})Xu, Sun, Zeng, Zhang, Ren, Wang, and
  Li}]{jingjing2018Upaired}
Jingjing Xu, Xu~Sun, Qi~Zeng, Xiaodong Zhang, Xuancheng Ren, Houfeng Wang, and
  Wenjie Li. 2018{\natexlab{a}}.
\newblock Unpaired sentiment-to-sentiment translation: {A} cycled reinforcement
  learning approach.
\newblock In \emph{Proceedings of the 56th Annual Meeting of the Association
  for Computational Linguistics, {ACL} 2018, Melbourne, Australia, July 15-20,
  2018, Volume 1: Long Papers}, pages 979--988. Association for Computational
  Linguistics.

\bibitem[{Xu et~al.(2018{\natexlab{b}})Xu, Zhang, Zeng, Ren, Cai, and
  Sun}]{xu2018emnlp}
Jingjing Xu, Yi~Zhang, Qi~Zeng, Xuancheng Ren, Xiaoyan Cai, and Xu~Sun.
  2018{\natexlab{b}}.
\newblock A skeleton-based model for promoting coherence among sentences in
  narrative story generation.
\newblock In \emph{Proceedings of the 2018 Conference on Empirical Methods in
  Natural Language Processing, {EMNLP} 2018, Brussels, Belgium, October
  31-November 4, 2018}. Association for Computational Linguistics.

\bibitem[{Xu et~al.(2015)Xu, Ba, Kiros, Cho, Courville, Salakhudinov, Zemel,
  and Bengio}]{xu2015show}
Kelvin Xu, Jimmy Ba, Ryan Kiros, Kyunghyun Cho, Aaron Courville, Ruslan
  Salakhudinov, Rich Zemel, and Yoshua Bengio. 2015.
\newblock Show, attend and tell: Neural image caption generation with visual
  attention.
\newblock In \emph{Proceedings of the 32nd International Conference on Machine
  Learning}, volume~37 of \emph{Proceedings of Machine Learning Research},
  pages 2048--2057, Lille, France. PMLR.

\bibitem[{Yao et~al.(2017)Yao, Pan, Li, Qiu, and Mei}]{yao2017boosting}
Ting Yao, Yingwei Pan, Yehao Li, Zhaofan Qiu, and Tao Mei. 2017.
\newblock Boosting image captioning with attributes.
\newblock In \emph{{IEEE} International Conference on Computer Vision, {ICCV}
  2017, Venice, Italy, October 22-29, 2017}, pages 4904--4912. {IEEE} Computer
  Society.

\bibitem[{You et~al.(2016)You, Jin, Wang, Fang, and Luo}]{you2016image}
Quanzeng You, Hailin Jin, Zhaowen Wang, Chen Fang, and Jiebo Luo. 2016.
\newblock Image captioning with semantic attention.
\newblock In \emph{2016 {IEEE} Conference on Computer Vision and Pattern
  Recognition, {CVPR} 2016, Las Vegas, NV, USA, June 27-30, 2016}, pages
  4651--4659. {IEEE} Computer Society.

\bibitem[{Young et~al.(2014)Young, Lai, Hodosh, and
  Hockenmaier}]{young2014image}
Peter Young, Alice Lai, Micah Hodosh, and Julia Hockenmaier. 2014.
\newblock From image descriptions to visual denotations: New similarity metrics
  for semantic inference over event descriptions.
\newblock \emph{Transactions of the Association for Computational Linguistics},
  2:67--78.

\bibitem[{Zhang et~al.(2006)Zhang, Platt, and Viola}]{zhang2006multiple}
Cha Zhang, John~C. Platt, and Paul~A. Viola. 2006.
\newblock Multiple instance boosting for object detection.
\newblock In Y.~Weiss, B.~Sch\"{o}lkopf, and J.~C. Platt, editors,
  \emph{Advances in Neural Information Processing Systems 18 [Neural
  Information Processing Systems, {NIPS} 2005, December 5-8, 2005, Vancouver,
  British Columbia, Canada]}, pages 1417--1424. MIT Press.

\bibitem[{Zhang et~al.(2017)Zhang, Kyaw, Chang, and Chua}]{zhang2017visual}
Hanwang Zhang, Zawlin Kyaw, Shih{-}Fu Chang, and Tat{-}Seng Chua. 2017.
\newblock Visual translation embedding network for visual relation detection.
\newblock In \emph{2017 {IEEE} Conference on Computer Vision and Pattern
  Recognition, {CVPR} 2017, Honolulu, HI, USA, July 21-26, 2017}, pages
  3107--3115. {IEEE} Computer Society.

\bibitem[{Zhao et~al.(2016)Zhao, Lu, Cai, He, and Zhuang}]{zhao2016partial}
Zhou Zhao, Hanqing Lu, Deng Cai, Xiaofei He, and Yueting Zhuang. 2016.
\newblock Partial multi-modal sparse coding via adaptive similarity structure
  regularization.
\newblock In \emph{Proceedings of the 2016 {ACM} Conference on Multimedia
  Conference, {MM} 2016, Amsterdam, The Netherlands, October 15-19, 2016},
  pages 152--156. {ACM}.

\end{thebibliography}
\bibliographystyle{acl_natbib_nourl}

\begin{table*}[t]
\centering
\scriptsize
\renewcommand\tabcolsep{4.5pt}
\begin{tabular}{@{}l c c c c c c c c c c c c c c c@{}}
\toprule
\multirow{2}{*}{COCO}
& \multicolumn{2}{c}{BLEU-1}     
&  \multicolumn{2}{c}{BLEU-2}     
&  \multicolumn{2}{c}{BLEU-3}    
&  \multicolumn{2}{c}{BLEU-4}   
&  \multicolumn{2}{c}{METEOR}  
& \multicolumn{2}{c}{ROUGE-L}
& \multicolumn{2}{c}{CIDEr}  \\
\cmidrule(lr){2-3} \cmidrule(lr){4-5} \cmidrule(lr){6-7} \cmidrule(lr){8-9} \cmidrule(lr){10-11} \cmidrule(lr){12-13} \cmidrule(lr){14-15}

& c5 & c40 & c5 & c40 & c5 & c40 & c5 & c40 & c5 & c40 & c5 & c40 & c5 & c40 \\
\midrule
HardAtt \cite{xu2015show}                      &0.705           & 0.881        & 0.528     & 0.779         & 0.383  &0.658 &0.277 &0.537 &0.241 &0.322 &0.516 &0.654 &0.865 &0.893     \\
ATT-FCN \cite{you2016image}                             &0.731           & 0.900         & 0.565     & 0.815         & 0.424 &0.709 &0.316 &0.599 &0.250 &0.335 &0.535 &0.682 &0.943 &0.958    \\
SCA-CNN \cite{chen2017scacnn}                           & 0.712         & 0.894    & 0.542     & 0.802    & 0.404 &0.691 &0.302 &0.579 &0.244 &0.331 &0.524 &0.674 &0.912 &0.921      \\
LSTM-A \cite{yao2017boosting}                           & 0.739     & 0.919     & 0.575    & 0.842     & 0.436 &0.740 &0.330 &0.632   &0.256 &0.350 &0.542 &0.700 &0.984 &1.003   \\
SCN-LSTM \cite{gan2017semantic}                         &0.740         & 0.917     & 0.575     &0.839         & 0.436 &0.739 &0.331 &0.631 &0.257 &0.348 &0.543 &0.696 &1.003 &1.013     \\
AdaAtt  \cite{lu2017knowing}\ssymbol{2}                            & 0.748     & 0.920     & 0.584    & 0.845     & 0.444 &0.744 &0.336 &0.637 &0.264 &0.359 &0.550 &0.705 &1.042 &1.059     \\
\midrule
TD-M-ATT \cite{chen2018temporal}\ssymbol{1}\ssymbol{2}             & 0.757        & 0.913     & 0.591     & 0.836     & 0.441 &0.726 &0.324 &0.609 &0.259 &0.342 &0.547 &0.689 &1.059 &1.090    \\
SCST \cite{rennie2017self}\ssymbol{1}\ssymbol{2}                   &0.781           & 0.937     & 0.619     & 0.860     & 0.470 &0.759 &0.352 &0.645 &0.270 &0.355 &0.563 &0.707 &1.147 &1.167     \\
Up-Down \cite{anderson2018bottom}\ssymbol{1}\ssymbol{2}\ssymbol{3} &\bf 0.802     & \bf 0.952 &\bf 0.641     & \bf0.888     & \bf 0.491 &\bf0.794 &\bf0.369 &\bf0.685 &\bf0.276 &\bf0.367 &\bf0.571 &\bf0.724 &\bf1.179 &\bf1.205 \\ \midrule
simNet                                                     & 0.766 & 0.941     &  0.605 & 0.874     & 0.462 &0.778 &0.350 &0.671 &0.267 &0.362 &0.558 &0.716 &1.087 &1.111     \\ \bottomrule
\end{tabular}
\caption{Performance on the online COCO evaluation server. The SPICE metric is unavailable for our model, thus not reported. c5 means evaluating against 5 references, and c40 means evaluating against 40 references. The symbol \ssymbol{1} denotes directly optimizing CIDEr. The symbol \ssymbol{2} denotes model ensemble. The symbol \ssymbol{3} denotes using extra data for training, thus not directly comparable. Our submission does not use the three aforementioned techniques. Nonetheless, our model is second only to Up-Down and surpasses almost all the other models in published work, especially when 40 references are considered.\label{tab:res-server}}
\end{table*}

\newpage

\appendix

\section{Supplementary Material}

\subsection{Results on COCO Evaluation Server}\label{sec:coco-server}
Table~\ref{tab:res-server} shows the performance on the online COCO evaluation server\footnote{\url{https://competitions.codalab.org/competitions/3221}}. We put it in the appendix because the results are incomplete and the SPICE metric is not available for our submission, which correlates the best with human evaluation. The SPICE metrics are only available at the leaderboard on the COCO dataset website\footnote{\url{http://cocodataset.org/\#captions-leaderboard}}, which, unfortunately, has not been updated for more than a year. Our submission does not directly optimize CIDEr, use model ensemble, or use extra training data. The three techniques typically result in orthogonal improvements \cite{lu2017knowing,rennie2017self,anderson2018bottom}. Moreover, the SPICE results are missing, in which the proposed model has the most advantage. Nonetheless, our model is second only to Up-Down \cite{anderson2018bottom} and surpasses almost all the other models in published work, especially when 40 references are considered.

\end{document}